\documentclass[preprint,5p]{elsarticle}

\usepackage{lineno,hyperref}
\modulolinenumbers[5]

\journal{Journal of Web Semantics}

\usepackage{numcompress}

\usepackage{paralist}
\usepackage{color}
\definecolor{grey}{RGB}{170,170,170}
\usepackage{array}
\usepackage{multirow}
\usepackage{bigstrut}
\usepackage{amsmath}
\usepackage{amsfonts}
\usepackage{enumitem}
\usepackage{scrextend}
\usepackage{subfig}
\hypersetup{
  pdfauthor = {Pavlos Vougiouklis},
  pdftitle = {Neural Wikipedian: Generating Textual Summaries from Knowledge Base Triples}
}

\makeatletter
\def\ps@pprintTitle{%
   \let\@oddhead\@empty
   \let\@evenhead\@empty
   \let\@oddfoot\@empty
   \let\@evenfoot\@oddfoot
}
\makeatother
\usepackage{changepage}

\newcolumntype{L}[1]{>{\raggedright\let\newline\\\arraybackslash\hspace{0pt}}m{#1}}
\newcolumntype{C}[1]{>{\centering\let\newline\\\arraybackslash\hspace{0pt}}m{#1}}
\newcolumntype{R}[1]{>{\raggedleft\let\newline\\\arraybackslash\hspace{0pt}}m{#1}}
\begin{document}

\begin{frontmatter}

\title{Neural Wikipedian: Generating Textual Summaries from Knowledge Base Triples}
\tnotetext[mytitlenote]{Fully documented templates are available in the elsarticle package on \href{http://www.ctan.org/tex-archive/macros/latex/contrib/elsarticle}{CTAN}.}

\author[soton]{Pavlos Vougiouklis}
\ead{pv1e13@ecs.soton.ac.uk}
\author[ujm]{Hady Elsahar}
\ead{hady.elsahar@univ-st-etienne.fr}
\author[soton]{Lucie-Aim\'{e}e Kaffee}
\author[ujm]{Christoph Gravier}
\ead{christophe.gravier@univ-st-etienne.fr}
\author[ujm]{Frederique Laforest}
\ead{frederique.laforest@univ-st-etienne.fr}
\ead{kaffee@soton.ac.uk}
\author[soton]{Jonathon Hare}
\ead{jsh2@ecs.soton.ac.uk}
\author[soton]{Elena Simperl}
\ead{e.simperl@soton.ac.uk}

\address[soton]{School of Electronics and Computer Science\\
  University of Southampton\\
  Southampton, United Kingdom}
\address[ujm]{Laboratoire Hubert Curien, CNRS\\
  UJM-Saint-\'{E}tienne\\
  Universit\'{e} de Lyon,\\
  Lyon, France}


\begin{abstract}
  Most people do not interact with Semantic Web data directly. Unless they have the expertise to understand the underlying technology, they need textual or visual interfaces to help them make sense of it. We explore the problem of generating natural language summaries for Semantic Web data. This is non-trivial, especially in an open-domain context. To address this problem, we explore the use of neural networks. Our system encodes the information from a set of triples into a vector of fixed dimensionality and generates a textual summary by conditioning the output on the encoded vector. We train and evaluate our models on two corpora of loosely aligned Wikipedia snippets and DBpedia and Wikidata triples with promising results.
\end{abstract}


\end{frontmatter}


\section{Introduction}
While Semantic Web data, such as triples in Resource Description Framework (RDF), is easily accessible by machines, it is difficult to be understood by people who are unfamiliar with the underlying technology. On the contrary, for humans, reading text is a much more accessible activity. In the context of the Semantic Web, Natural Language Generation (NLG) is concerned with the implementation of textual interfaces that would effectively increase humans' accessibility to the information that is stored in the knowledge bases' triples. Further development of systems for NLG could be beneficial in a great range of application domains. Examples include Question Answering platforms whose users' experience could be improved by the ability to automatically generate a textual description of an entity that is returned at a user's query (e.g. the Google Knowledge Graph\footnote{\href{https://googleblog.blogspot.co.uk/2012/05/introducing-knowledge-graph-things-not.html}{\texttt{https://googleblog.blogspot.co.uk}}} and the Wikidata Reasonator\footnote{\href{https://tools.wmflabs.org/reasonator}{\texttt{https://tools.wmflabs.org/reasonator}}}), or dialogue systems in commercial environments that could be enhanced further in order to generate responses that better address the users' questions \cite{Janzen2009}.

So far, research has mostly focused on adapting rule-based approaches to generate text from Semantic Web data. These systems worked in domains with small vocabularies and restricted linguistic variability, such as football match summaries \cite{Bouayad-Agha2012} and museum exhibits' descriptions \cite{Dannells2012}. However, the tedious repetition of their textual patterns along with the difficulty of transferring the involved rules across different domains or languages prevented them from becoming widely accepted \cite{Bouayad-Agha2014}.

We address the above limitations by proposing a statistical model for NLG using neural networks. Our work explores how an adaptation of the encoder-decoder framework \cite{Cho2014,Sutskever2014} could be used to generate textual summaries for triples. More specifically, given a set of triples about an entity (i.e. the entity appears as the subject or the object of the triples), our task consists in summarising them in the form of comprehensible text. We propose a model that consists of a feed-forward architecture that encodes each triple from an input set of triples in a vector of fixed dimensionality in a continuous semantic space, and an RNN-based decoder that generates the textual summary one word at a time. Our model jointly learns unique vector representations ``embeddings'' for entities and words that exist in the text, and predicates and entities as they occur in the corresponding triples. In contrast with less flexible, rule-based strategies for NLG, our approach does not constrain the number of potential relations between the triples' predicates and the generated text. Consequently, a learnt predicate embedding, given its position in the semantic space, can be expressed in an varied number of different ways in the text.

Training data for NLG models is not always readily available; this applies to the context of Semantic Web as well. The difficulty is that data that is available in knowledge bases needs to be aligned with the corresponding texts. Existing solutions for data-to-text generation either focus mainly on creating a small, domain-specific corpus where data and text are manually aligned by a small group of experts, such as the WeatherGov \cite{Liang2009} and RoboCup \cite{Chen2008} datasets, or rely heavily on crowdsourcing \cite{Mrabet2016}, which makes them costly to apply for large domains. Our second contribution is an automatic approach for building a large data-to-text corpus of rich linguistic variability. We rely on the alignment of DBpedia and Wikidata with Wikipedia in order to create two corpora of knowledge base triples from DBpedia and Wikidata, and their corresponding textual summaries. For the purpose of this paper, we chose to retrieve articles about people and their biographies \cite{Lebret2016}. We extracted two different corpora with vocabularies of over $400$k words that consist of: \begin{inparaenum}[(i)]\item $260$k Wikipedia summaries aligned with a total of $2.7$M DBpedia triples, and \item $360$k Wikipedia summaries allocated to a total of $43$M Wikidata triples\end{inparaenum}.

Our proposed model learns to generate a textual summary as a sequence of words and entities. We experiment with two different approaches, one rule-based, and one statistical, in order to infer the verbalisation of the predicted entities in the generated summary. Conventional systems based on neural networks when employed on NLG tasks, such as Machine Translation \cite{Sutskever2014} or Question Generation \cite{Serban2016} are incapable of learning high quality vector representation for the infrequent tokens (i.e. either words or entities) in their training dataset. Inspired by \cite{Luong2015a,Serban2016}, we address this problem by adapting a multi-placeholder method that enables the model to emit special tokens that map a rare entity in the text to its corresponding triple in the input set. We use perplexity, and the BLEU and ROUGE metrics in order to automatically evaluate our approach's ability of predicting the Wikipedia summary that corresponds to a set of unknown triples showing substantial improvement over our baselines. Furthermore, we evaluate a set of generated summaries against human evaluation criteria. Based on the average rating across our selected criteria, we conclude that our approach is able to generate coherent textual summaries that address most of the information that is encoded in the input triples. Lastly, we demonstrate our method's capability to successfully infer semantic relationships among entities by computing the nearest neighbours of the learned embeddings of respective entities in our datasets.

The structure of the paper is as follows. Section \ref{sec:Related} discusses existing approaches to NLG and the Semantic Web, and relates them to our model. Section \ref{sec:Model} presents the components of our approach. Section \ref{sec:Dataset} describes the construction of our datasets. Section \ref{sec:Experiments} presents experiments and an evaluation of the model. Section \ref{sec:Conclusion} summarises the contributions of this work and outlines future plans.

\section{Related Work}
\label{sec:Related}

Models for NLG can be divided into two groups: statistical and rule-based ones \cite{Reiter2000}. The latter employ linguistic expertise and work in three different phases:\begin{inparaenum}[(i)]\item document planning or content selection, \item microplanning and \item surface realisation \cite{Reiter2000,Bouayad-Agha2014}\end{inparaenum}. During document planning the information that will be communicated in the text is selected and organised (i.e. document structuring). The output of the document planner is used by the microplanner to decide how this information should be linguistically expressed in the generated text. Subsequently, the realiser generates the actual text by applying a specific template that satisfies the linguistic requirements that are set by the microplanner, and expresses the information as it is structured by the document planner. Each one of the above mentioned stages is associated almost explicitly not only with the domain of the end-application but, in most cases, with the application itself.

Most of the previous work on NLG with Semantic Web data has focused on the verbalisation of domain ontologies by using rules. Examples include systems that generate text in domains with limited linguistic variability, such as clinical narratives \cite{Arguello2011}, summaries of football matches \cite{Bouayad-Agha2012}, and, descriptions of museum's exhibits \cite{Dannells2012}. Further Semantic Web oriented NLG applications can be found in \cite{Bouayad-Agha2014}. Our work naturally lies on the path opened by recent unsupervised \cite{Duma2013} and \textit{distant-supervision} \cite{Ell2014} based approaches for the extraction of RDF verbalisation templates using parallel data-to-text corpora. However, rather than making a prediction about the template that would be the most appropriate to verbalise a set of input triples, our model jointly performs content selection and surface realisation, without the inclusion of any hand-coded rules or templates.

Previous work on neural network approaches shows their great potential at tackling a wide variety of NLP tasks ranging from machine translation \cite{Cho2014,Sutskever2014} to automatic response generation \cite{Vinyals2015,Vougiouklis2016}, and to computing vector representations of words in a continuous semantic space \cite{Mikolov2013a}. Our approach is inspired by the general encoder-decoder framework \cite{Cho2014,Sutskever2014} with multi-gated Recurrent Neural Network (RNN) variants, such as the Gated Recurrent Unit (GRU) \cite{Cho2014} and the Long Short-Term Memory (LSTM) cell \cite{Hochreiter1997}. Adaptations of this framework have demonstrated state-of-the-art performance in many generative tasks, such as machine translation \cite{Cho2014,Sutskever2014,Vinyals2015b}, and conversation modelling and response generation \cite{Wen2015,Vinyals2015}.

Implementations based on the encoder-decoder framework work by mapping sequences of source tokens to sequences of target tokens. We adapt the Se\-quence-to-Se\-quence model to the requirements of Semantic Web Data. Since sets of triples are unordered, and not sequentially correlated, in the next section we propose a model that consists of a feed-forward neural network that encodes each input triple into a vector of fixed dimensionality in a continuous semantic space. Within this space, triples that have similar semantic meaning will have similar positions. We couple this novel encoder with an RNN-based decoder that generates the textual summary one token (i.e. a token can be a word or an entity or a surface form of an entity) at a time.

Our task is most similar to recent work by \citeauthor{Lebret2016} and \citeauthor{Chisholm2017}, who both employ adaptations of the encoder-decoder framework in order to generate the first sentence of a Wikipedia biography \cite{Lebret2016,Chisholm2017}. \citeauthor{Lebret2016} propose a system that it generates a summary given an input in the form of a Wikipedia infobox; the model proposed by Chisholm generates a biography given a sequence of slot-value pairs extracted from Wikidata. The representation of the input of both these approaches is essentially limited to expressing only one-subject relationships. In our case, the input triples set that is allocated to each Wikipedia summary is made of more than just the DBpedia or Wikidata triples of the corresponding Wikipedia article. As we note in more detail in Section \ref{subsec:Triples}, this triple set also includes triples with entities that are related with the main discussed entity of a Wikipedia biography in the respective knowledge base, and their object is the main subject of the Wikipedia summary. Furthermore, we believe that constraining the generative process to only the first sentence significantly simplifies the task in terms of the amount of information (i.e. in our case number of triples) that is lexicalised in the summary. Consequently, we choose to train on longer snippets of text to generate more elaborate summaries.

\section{Our Model}
\label{sec:Model}

\begin{table}[t]
  \caption{An idealised example of our NLG task. Our system takes as an input a set of triples about \textit{Walt Disney}, whose either subject or object is related to the entity of Walt Disney, and it generates a textual summary.}
  \begin{center}
    \scriptsize
    \setlength{\extrarowheight}{1.5pt}
    \begin{tabular}{|R{1.5cm}|L{6.4cm}|}\hline
      \multirow{3}{*}{\textbf{Triples}} & \texttt{dbr:Walt\_Disney dbo:birthDate ``1901-12-05''} \\
                                        & \texttt{dbr:Walt\_Disney dbo:birthPlace dbr:Chicago} \\
                                        & \texttt{dbr:Mickey\_Mouse dbo:creator dbr:Walt\_Disney} \\ \hline
      \textbf{Textual Summary} & Walt Disney was born in Chicago, and was the creator of Mickey Mouse. \\ \hline       
    \end{tabular}
    
    \label{table:SimpleTextGenerationExample}
  \end{center}
\end{table}

An idealised example of our NLG task is presented in Table \ref{table:SimpleTextGenerationExample}; our system takes as an input a set of triples about the entity \textit{Walt Disney} (i.e. the entity \textit{Walt Disney} is either the subject or object of the triples in the set), and generates a sequence of words in order to summarise them in the form of comprehensible text. Given a set of $E$ triples, $F = \{f_1, f_2, \ldots, f_E\}$, our goal is to learn a model that is able to generate a sequence of $T$ tokens, $Y=y_1, y_2, \ldots, y_{T}$. We regard $Y$ as a representation in natural language of the input set of triples, and we build a model that computes the probability of generating $y_1, y_2, \ldots, y_{T}$, given the initial set of triples $f_1, f_2, \ldots, f_E$:
\begin{align}
  p(y_1, \ldots, y_{T} | f_1, f_2, \ldots, f_E) = \prod_{t=1}^{T}p(y_t|y_1, \ldots y_{t-1}, F) \enspace.
\end{align}

\begin{figure}[h]
  \centering
  
  \def\svgwidth{.985\linewidth}
  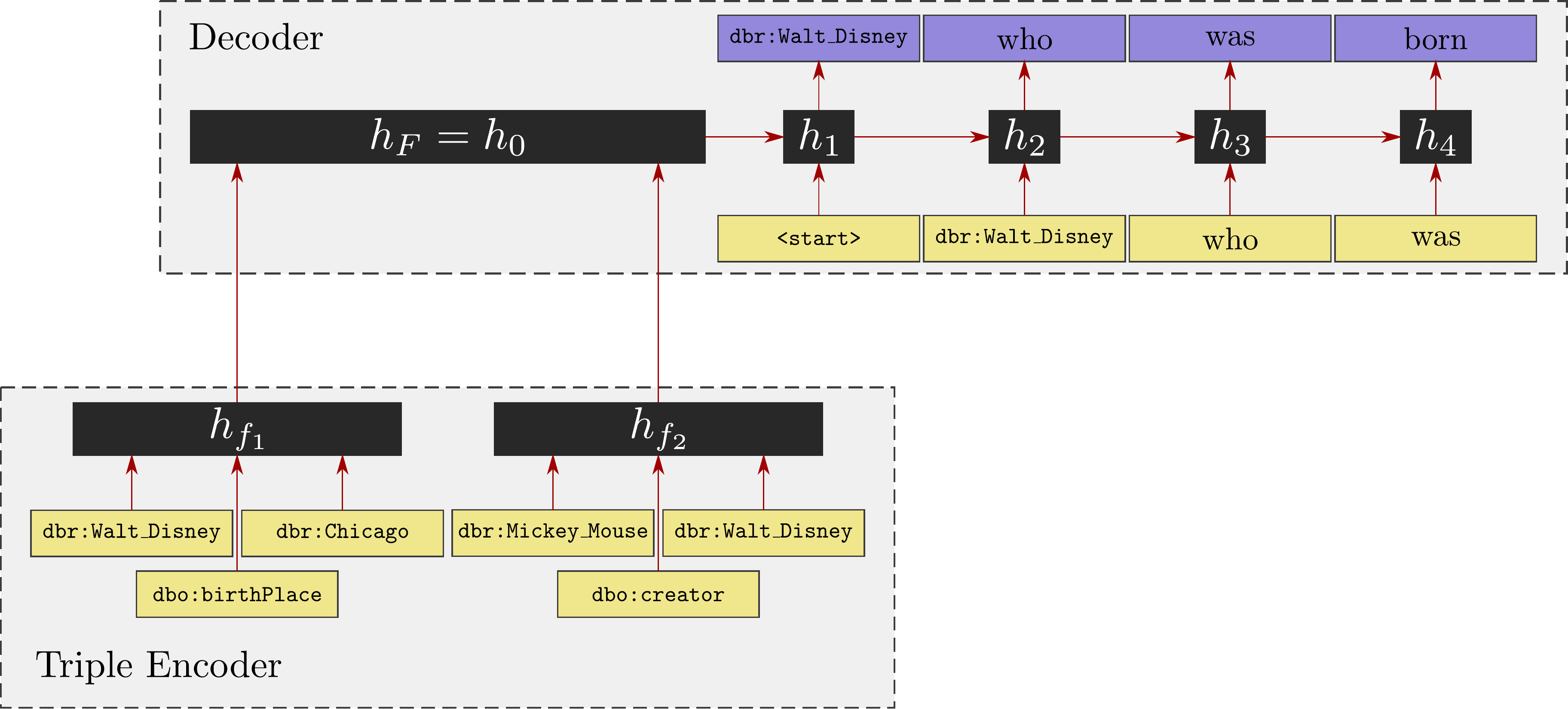
  \caption{The triple encoder computes a vector representation for each one of the two input triples $h_{f_1}$ and $h_{f_2}$. Subsequently, the decoder uses the concatenation of the two vectors, $[h_{f_1};h_{f_2}]$ to initialise the decoding process that generates the summary word per word. Each textual summary starts and ends with the respective start-of-sequence {\tt<start>} and end-of-sequence {\tt<end>} tokens.}
  \label{fig:SequentialArchitecture}
\end{figure}

Our model consists of a feed-forward architecture that encodes each triple from the input set into a vector of fixed dimensionality in a continuous semantic space. This is coupled to an RNN-based decoder that generates the textual summary one token (i.e. a token can be a word or an entity or a surface form of an entity) at a time. Note that since \textit{bias} terms can be included in each weight-matrix multiplication, they are not explicitly shown in the equations that describe the models of this section. The architecture of our generative model is shown in Figure \ref{fig:SequentialArchitecture}.

\subsection{Triple Encoder}

Let $F = \{f_1,f_2, \ldots, f_E: f_i=(s_i,p_i,o_i)\}$ be the set of triples $f_1, \ldots, f_E$, where $s_i$, $p_i$ and $o_i$ are the one-hot\footnote{One-hot is a vector that contains a $1$ at the index of this particular $s_i$, $p_i$ and $o_i$ token in the vocabulary with all the other values set to zero.} vector representations of the respective subject, predicate and object of the $i$-th statement. The vector representation $h_{f_i}$ of the $i$-th triple is computed by forward propagating the triples encoder as follows:

\begin{align}
  \widetilde{h_{f_i}} & =  [\mathbf{W_{x \rightarrow \widetilde{h}}}s_i;\mathbf{W_{x \rightarrow \widetilde{h}}}p_i;\mathbf{W_{x \rightarrow \widetilde{h}}}o_i] \enspace, \\
  h_{f_i} & = \text{ReLU}(\mathbf{W_{\widetilde{h} \rightarrow h}}\widetilde{h_{f_i}}) \enspace,
\end{align} where $\text{ReLU}$ is the rectifier (i.e. non-linear activation function), $[\ldots;\ldots]$ represents vector concatenation, $\mathbf{W_{x \rightarrow h}}:\mathbb{R}^{|N|} \rightarrow \mathbb{R}^m$ is a trainable weight matrix that represents a biased linear mapping, where $|N|$ is the cardinality of all the potential one-hot input vectors (i.e. size of the dictionary of all the available predicates and entities of the triples dictionary), and $\mathbf{W_{\widetilde{h} \rightarrow h}}:\mathbb{R}^{3m} \rightarrow \mathbb{R}^m$ is an unbiased linear mapping.

\subsection{Decoder}
After the vector representation $h_{f_i}$ for each triple $f_i$ is obtained, we start the decoding process during which the corresponding textual summary is generated. At each timestep $t$ the decoder makes a prediction about the next token that will be appended to the summary by taking into consideration both the tokens that have already been generated, and the contextual knowledge from the triples that have been provided to the system initially as input. We experiment with two commonly used RNN variants:\begin{inparaenum}[(i)]\item the LSTM cell and \item the GRU\end{inparaenum}, in order to explore which one works best for the decoding needs of our architecture.

We initialise the decoder with a fixed-length vector that we obtain after encoding all the information from the vector representations of the triples. Our approach is inspired by the general Sequence-to-Sequence framework, within which an RNN-based encoder encapsulates the information that exists in a sequence, and an RNN-based decoder that generates a new sequence from this encapsulation \cite{Cho2014,Sutskever2014}. However, since the triples that we use in our problem are not sequentially correlated, we propose a concatenation-based formulation that enables us to capture the information across all the triples that are given as an input to our system into one single vector. More specifically, given a set of triples' vector representations, $h_{f_1}, \ldots, h_{f_E}$, we compute:

\begin{align}
  \widetilde{h_F} & = [h_{f_1}; h_{f_2}; \ldots; h_{f_{E-1}}; h_{f_E}] \enspace, \\
  h_F & = \mathbf{W_{h_F \rightarrow h_0^1}}\widetilde{h_F} \enspace,
\end{align}where $\mathbf{W_{h_F \rightarrow h_0^1}}:\mathbb{R}^{Em} \rightarrow \mathbb{R}^m$ is a biased linear mapping. Subsequently, the hidden units of the LSTM- or GRU-based decoder (discussed below) at layer depth $l=1$ are initialised with $h_0^1=h_F$.

Let $h_t^l \in \mathbb{R}^{m}$ be the aggregated output of a hidden unit at timestep $t=1... T$ and layer depth $l=1... L$. The vectors at zero layer depth, $h_t^0=\mathbf{W_{x \rightarrow h}}x_t$, represent the words or entities that are given to the network as an input. The parameter matrix $\mathbf{W_{x \rightarrow h}}$ has dimensions $[|X|, m]$, where $|X|$ is the cardinality of all the potential one-hot input vectors (i.e. size of the dictionary of all the available words and entities of the textual summaries dictionary). All subsequent matrices have dimension $[m, m]$ unless stated otherwise.

\subsubsection{Long Short-Term Memory (LSTM).}
We adopt the architecture from \cite{Zaremba2014}:

\begin{align} 
  \left (
    \begin{array}{l}
      in_t^l \\
      f_t^l \\
      out_t^l \\
      \widetilde{c_t^l}
    \end{array}
  \right ) & = \left (
    \begin{array}{l}
      \text{sigm} \\
      \text{sigm} \\
      \text{sigm} \\
      \text{sigm}
    \end{array}
  \right ) \mathbf{W^l} \left (
    \begin{array}{l}
      h_{t}^{l-1}\\
      h_{t-1}^{l}
    \end{array}
  \right ) \enspace,\\
  c_t^l & = f_t^l \odot c_{t-1}^l + in_t^l \odot \widetilde{c_t^l} \enspace,\\
  h_t^l & = out_t^l \odot \text{tanh}(c_t^l)\enspace,
\end{align}where $\mathbf{W^l}:\mathbb{R}^{4m} \rightarrow \mathbb{R}^{2m}$ is a biased linear mapping, and $in_t^l$, $f_t^l$, $out_t^l$ and $c_t^l$ are the vectors at timestep $t$ and layer depth $l$ that correspond to the \textit{input gate}, the \textit{forget gate}, the \textit{output gate} and the \textit{cell} respectively.

\subsubsection{Gated Recurrent Unit (GRU).}
The GRU is a less complex variant of the LSTM cell \cite{Cho2014} with comparable performance \cite{Chung2014}.

\begin{align} 
  \left (
    \begin{array}{l}
      r_t^l \\
      u_t^l \\
    \end{array}
  \right ) &= \left (
    \begin{array}{l}
      \text{sigm} \\
      \text{sigm} \\
    \end{array}
  \right ) \mathbf{W^l} \left (
    \begin{array}{l}
      h_{t}^{l-1}\\
      h_{t-1}^{l}
    \end{array}
  \right ) \enspace,\\
  \widetilde{h_t^l} & = \text{tanh}(\mathbf{W_{in}^l}h_{t}^{l-1} + \mathbf{W_{h \rightarrow h}^l} (r_t^l \odot h_{t-1}^{l}))\enspace,\\
  h_t^l & = (1 - u_t^l) \odot h_{t-1}^{l} + u_t^l \odot \widetilde{h_t^l}\enspace,
\end{align}where $\mathbf{W^l}:\mathbb{R}^{2m} \rightarrow \mathbb{R}^{2m}$ is a biased linear mapping, and $r_t^l$, $u_t^l$ and $\widetilde{h_t^l}$ are the vectors at timestep $t$ and layer depth $l$ that represent the values of the \textit{reset gate}, the \textit{update gate} and the \textit{hidden candidate} respectively.

\subsection{Model Training}
\label{subsec:ModelTraining}

The conditional probability distribution over the each token of the summary at each timestep $t$ is represented with the softmax function over all the entries in the textual summaries dictionary:

\begin{align} \label{eq:softmax}
  p(y_t|y_1, y_2, \ldots y_{t-1}, F) = \text{softmax}(\mathbf{W_y}h_t^L) \enspace,
\end{align}where $\mathbf{W_y}:\mathbb{R}^{m} \rightarrow \mathbb{R}^{|X|}$ is a biased trainable weight matrix. Our model learns to make a prediction about the next token by using the negative cross-entropy\footnote{In information theory, entropy $H$ is a measure of the uncertainty. The concept of cross-entropy is associated with the similarity between two distributions, an empirical one $q$ and a predicted one $p$ given a random variable $X$ and set of parameters $\theta$. It is defined as: $H(X)= -\sum q(y^{(i)})\log p(y^{(i)}|x^{(i)}, \theta)$.} criterion. During training and given a set of triples, our model makes a prediction about the sequence of tokens of which the generated summary is comprised. The model computes how far the generated sequence of tokens is from the empirical, actual text by utilising the negative logarithmic probability of the generated summary given set of triples:
\begin{align}
  \text{cost} = - \sum_{t=1}^{T} \log\,p(y_t|y_1, y_2, \ldots y_{t-1}, F)\enspace.
\end{align}Consequently, our model tries to minimise the above cost function. This non-convex optimisation problem is solved using the RMSProp\footnote{RMSProp stands for Root Mean Square Propagation, and is a form of stochastic gradient descent where the gradient for each weight is divided by a running average of its recent gradients norm \cite{Tieleman2012}.} algorithm.

\subsection{Generating Summaries}
\label{subsec:GeneratingSummary}

\begin{figure*}[h]
  \centering
  
  \def\svgwidth{.95\linewidth}
  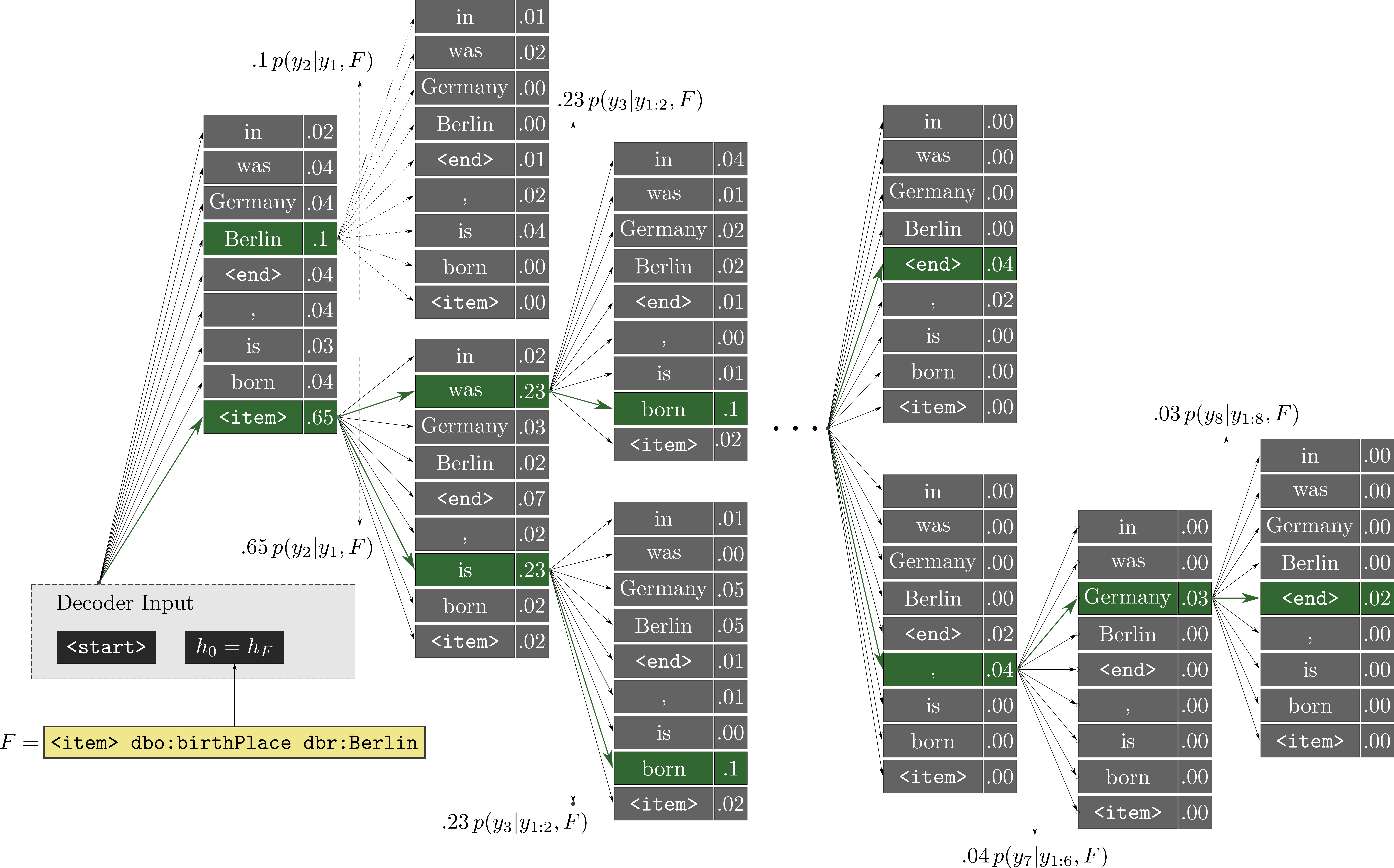
  \caption{An idealised example of a beam-search decoder with a beam $B$ of size $2$ and target vocabulary size $|X|$ equal to $9$. The scores at the right-hand side of the words in the vocabulary is the probability of the summary when it is extended by that particular word.}
  \label{fig:beam-search}
\end{figure*}

During testing, our goal is to find:
\begin{align}
  \mathbf{y^*} = \operatornamewithlimits{\arg \max}\limits_{\mathbf{y}} \sum_{t=1}^{T} \log\,p(y_t|y_1, \ldots y_{t-1}, F)\enspace,
\end{align}where $\mathbf{y^*}$ is the optimal summary computed by the model. Recall from Eq. \ref{eq:softmax} that at each timestep, our model predicts a probability distribution over the token that is more likely to come next. In theory, Viterbi decoding could approximate an optimal summary. However, in practice, the fact that the target vocabulary $|X|$ is large enough deems such an approach intractable. A different approach is to approximate the best summary by appending the token with the highest probability at each timestep of the generation process. Even though, such greedy decoders have proven to be very fast when employed in machine translation problems, they tend to produce low quality approximations \cite{Rush2015}.

A compromise between a strictly-greedy decoding algorithm and Viterbi is to adopt a beam-search decoder \cite{Sutskever2014}, which provides us with the $B$-most-probable summaries or hypotheses given a set of triples $F$ as input. The decoder maintains only a small number of $B$ hypotheses (i.e. partially completed summaries) and which it extends at every timestep with every token in the target vocabulary $|X|$.

During testing, we provide our network with an unknown set of triples, and we initialise the decoder with a special start-of-summary \texttt{<start>} token. The $B$ tokens with the highest probability are used as separate inputs to the decoder at the second timestep. This leads to $B|X|$ partial hypotheses from which we only retain the $B$-best. After all the second words of our hypotheses are provided as an input to the decoder, we end up with $B|X|$ partial three-worded hypotheses from which again we only keep the $B$ ones with the highest probability. When the end-of-summary \texttt{<end>} token is predicted for a hypothesis, it is appended to the list of complete summaries, and the process carries on with $B=B-1$. An idealised example of a beam-search decoder is displayed in Figure \ref{fig:beam-search}.

\section{Datasets}
\label{sec:Dataset}

In order to address the training needs of our proposed model, we build two datasets of aligned knowledge base triples with texts. For the first dataset, we leverage the intrinsic alignment of DBpedia and Wikipedia in order to create a corpus of loosely aligned triples and textual summaries. In our second scenario, following [Elsahar et al. 2017], we align Wikipedia summaries with the community-curated triples of Wikidata.

Inspired by \citeauthor{Lebret2016} , we chose a corpus about biographies. Biographies represent one of the largest single domains in Wikipedia, providing us with a substantial amount of training data [cite?]. We believe they offer the necessary diversity and linguistic variability that will allow us to explore effectively the generative ability of our systems. In addition, the fact that the biographies tend to adopt a limited number of structural paradigms, provides us with the opportunity to better understand the limitations of our approach.

By using PetScan\footnote{PetScan (i.e. \href{https://petscan.wmflabs.org/}{\texttt{petscan.wmflabs.org}}) is a tool that identifies Wikipedia articles, images and categories based on the category or subcategory to which they belong.}, we collected a list of $1,479,170$ Wikipedia articles that have been detected by the WikiProject Biography\footnote{\href{https://en.wikipedia.org/wiki/Wikipedia:WikiProject_Biography}{\texttt{en.wikipedia.org/wiki/Wikipedia:WikiProject\_Biography}}}. We then extracted the Wikipedia summaries and their corresponding DBpedia triples from the Mapping-based Objects\footnote{\label{dbpedia.org}\href{http://wiki.dbpedia.org/downloads-2016-10}{\texttt{http://wiki.dbpedia.org/downloads-2016-10}}} and Literals\footref{dbpedia.org} DBpedia dataset, retaining only articles for which an infobox exists. For the Wikidata version of the dataset, we used the Wikidata truthy dumps \footnote{\href{https://dumps.wikimedia.org/wikidatawiki/entities}{\texttt{https://dumps.wikimedia.org/wikidatawiki/entities}}} and we kept only items for which Wikidata triples exist.

In addition to the above datasets, we also leverage two DBpedia datasets:\begin{inparaenum}[(i)]\item the Instance Types\footref{dbpedia.org} and \item the Genders\footref{dbpedia.org} datasets\end{inparaenum}. The first one is used to provide us with special tokens for the entities that occur infrequently in our aligned datasets, and the second in order for us to append a gender-related triple to the DBpedia triples that have been already allocated to an article. Since co-reference resolution is not performed as a data pre-processing stage, our hypothesis is that the additional knowledge from the inclusion of gender-related triples will increase the model's awareness towards the gender of the main discussed entity of an article. Please note that the Genders dataset is used for the DBpedia version of the aligned dataset, in which gender-related triples are extremely sparse.

\subsection{Wikipedia Summaries}
\label{subsec:WikipediaSummaries}
  
One of the main challenges that is associated with the alignment of triples from a structured knowledge base with texts is the identification of how the entities of the knowledge base are mentioned in the text. For instance in the Wikipedia sentence: ``Barack Hussein Obama II is an American politician who served as the 44th President of the United States from 2009 to 2017.''\footnote{\href{https://en.wikipedia.org/wiki/Barack_Obama}{\texttt{https://en.wikipedia.org/wiki/Barack\_Obama}}} we need to be able to identify that the surface forms of ``Barack Hussein Obama II'' and ``United States'' refer to the respective DBpedia resources of \texttt{dbr:Barack\_Obama} and \texttt{dbr:United\_\-States}. In order to sidestep this problem, we use DBpedia Spotlight~\cite{Daiber2013}, an automatic system for annotation of DBpedia entities in text. \textit{Confidence} and \textit{support} are the two main variables that parameterise the annotation results that are returned by DBpedia Spotlight. The first one is the lowest threshold of certainty which the system must have in order to return an annotation, and the latter is the lowest bound of the un-normalised total number of links to the returned entities.

We run each one of the extracted Wikipedia summaries through DBpedia Spotlight. Our goal was to find the combination that provides the greatest number of relevant annotations, in order to\begin{inparaenum}[(i)]\item enhance the set of triples allocated to each Wikipedia page more effectively, and to \item allow the model to learn directly how entities in the triples on the encoder side manifest themselves in the text on the decoder side\end{inparaenum}. We empirically found that by setting the \textit{confidence} and \textit{support} parameters to $0.35$ and $-1$ respectively, we increased the recall of the identified resources while maintaining the precision at acceptable levels. We retained a list of all the possible surface forms to which each entity was mapped. Furthermore, given the nature of our problem, we excluded any Wikipedia summaries whose main discussed entity was not identified in the text.

Each Wikipedia summary is tokenised and split into sentences using the Natural Language Toolkit (NLTK) \cite{Bird2009}.
  
\subsection{Knowledge Base Triples}
\label{subsec:Triples}

Our text generation task consists of learning how entities, along with their relationships, are mentioned in the text. Given a set of triples, our approach learns to generate text one token at a time, without constraining the generative procedure to pre-defined templates that would include a given textual string \textit{as-it-is} in the generated summary. Consequently, we excluded from our corpus any triples with a textual string as their object, except those that referred to numbers, dates or years. All instances of number-objects are replaced with the special token \texttt{0}, except for year-objects that are mapped to the special \texttt{<year>} token \cite{Lebret2016}. In both Wikidata and DBpedia date-related objects are expressed as a string followed by its corresponding XML Schema URI (e.g. \texttt{XMLSchema/\-\#dateTime} or \texttt{XMLSchema/\-\#date}). In order to enable our model to process date-related triples and learn how their information is lexicalised in the text, we decompose them into two different triples. The first one is used to represent the month as it has been identified in the original triple, and the second one to represent the year. The object of the latter is subsequently mapped to the special \texttt{<year>} token. Table \ref{table:DateEncoding} presents an example of our date encoding approach.

\begin{table}[t]
  \caption{An example of how a triple whose object is identified as a date is encoded into two different triples. The first one represents the month that has been identified in the original triple, and the second the year.}
  \begin{center}
    \scriptsize
    \setlength{\extrarowheight}{1.5pt}
    \begin{tabular}{|R{2.05cm}|L{5.85cm}|}\hline
      \textbf{Original Triple} & \texttt{dbr:Andre\_Agassi dbo:birthDate ``1970-04-29''} \\ \hline
      \multirow{2}{2cm}{\raggedleft \textbf{Resultant Triples}} & \texttt{dbr:Andre\_Agassi dbo:birthDateMonth 4} \\
                                         & \texttt{dbr:Andre\_Agassi dbo:birthDateYear <year>} \\ \hline
    \end{tabular}
    \label{table:DateEncoding}
  \end{center}
\end{table}

For each entity that has been identified in a Wikipedia summary using DBpedia Spotlight, we extracted its corresponding triples from the Mapping-based Objects dataset in the DBpedia's case, and the Wikidata truthy dump in Wikidata's case. We assume that the subjects or objects of a set of triples are consistent with the main subject of the corresponding Wikipedia summary. Consequently, from this additional set of triples we only retain those whose object matches the main discussed entity in each summary, and we append them to the initial set. This results in $450$ and $609$k unique predicates and entities in DBpedia's case and in $378$ and $378$k unique predicates and entities respectively in Wikidata's case.

\subsection{Aligned Dataset}
\label{subsec:AlignedDataset}

We built two aligned datasets that consist of:\begin{inparaenum}[(i)]\item $256850$ instances of Wikipedia summaries aligned with $2.74$M DBpedia triples, and \item $358908$ instances of Wikipedia summaries aligned with the total of $4.34$M Wikidata triples respectively\end{inparaenum}. The size difference of our datasets is explained in a twofold manner. Firstly, there are Wikipedia biographies without an infobox (i.e. and, thus, without any available triples in the Mapping-based Objects and Literals DBpedia datasets). Secondly, even if they do have an infobox, the retrieved triples that are made available in the DBpedia dumps might not meet the requirements of our task (i.e. Section \ref{subsec:Triples}). For example in case the objects of all the triples that allocated to a Wikipedia biography are strings other than dates or number then this summary would be excluded from the respective aligned dataset.

We describe next all the pre-processing steps that we followed in order to make our aligned datasets fit for the training our neural network architectures.

\subsubsection{Modelling the Generated Summaries}
\label{subsubsec:ModellingGeneratedSummaries}

We retained only the first two sentences of each summary in order to reduce the computational cost of our task; summaries that consist of only one sentence were included unaltered. Since it would be impossible to learn a unique vector representation for the main discussed entity of each Wikipedia summary due to the lack of occurrences of the majority of those entities in the datasets, we replaced them with the special \texttt{<item>} token. We used a fixed vocabulary of $30000$ and $32000$ of the most frequent tokens (i.e. either words or entities) of the summaries that are aligned with the respective DBpedia and Wikidata triples. Similarly to the input triples (i.e. Section \ref{subsec:Triples}), all occurrences of numbers in the text are replaced with the special token \texttt{0}, except for year-objects that are mapped to the special \texttt{<year>} token \cite{Lebret2016}. Every out-of-vocabulary word is represented by the special \texttt{<rare>} token.

Using a single special token for all the rare entities that have not been included in the fixed target vocabulary would substantially limit the model, causing unnecessary repetition of this particular token in the generated summaries. Inspired by the Multi-Placeholder model \cite{Serban2016}, we first attempt to match a rare entity that has been annotated in the text, in the subjects or the objects of the allocated triples. In case it exists in the triples, then it is replaced by a placeholder token, which consists from the predicate of the triple, a descriptor of the component of the triple that was matched (i.e. \texttt{\_\_obj\_\_} or \texttt{\_\_subj\_\_}), and the instance type of the entity. The instance type of an entity is obtained from the Instance Types dataset. For example, in the case that the subject of the triple: (\texttt{dbr:The\_Adventures\_of\-\_Roderick\_Random} \texttt{dbo:author} \texttt{dbr:Tobias\_Smollett}) is annotated as a rare entity in the corresponding summary, it is replaced with the special token: \texttt{dbo:author\_\_sub\_\_\-dbo:Book}. In the case that a rare entity is matched to the object of the triple: (\texttt{Kevin\_Carr} \texttt{dbo:birthPlace} \texttt{dbr:Morpeth,\_Northumbe\-rland}) it is replaced with the token: \texttt{dbo:birthPlace\_\-\_obj\_\_dbo:Settlement}. We refer to those placeholders as property-type placeholders. In case the entity does not have a type in the Instance Types dataset, the instance type part of the placeholder is filled by \texttt{<unk>} token (e.g. \texttt{dbo:birthPlace\_\_obj\_\_<unk>}). If the rare entity is not matched to any subject or object of the set of corresponding triples, then it is replaced by the special token of its instance type. In case the rare entity does not exist in the instance types dataset, it is replaced by the \texttt{<unk>} token.

Note that each summary is augmented with the respective start-of-summary \texttt{<start>} and end-of-summary \texttt{<end>} tokens.

\begin{table*}[h]
  \caption{Statistics regarding the initial and the training version of our two corpora based on DBpedia and Wikidata triples.}
  \begin{center}
    \footnotesize
    \setlength{\extrarowheight}{1.5pt}
    \begin{tabular}{|R{4.5cm}|C{2.5cm}|C{2.5cm}||C{2.5cm}|C{2.5cm}|}
      \hline
      \multirow{2}{*}{\textbf{Parameter}} & \multicolumn{2}{c||}{\textbf{DBpedia}} & \multicolumn{2}{c|}{\textbf{Wikidata}} \\ \cline{2-5}
                                          & Initial Dataset & Training Dataset & Initial Dataset & Training Dataset \\ \hline
      Total $\#$ of Articles & $256850$ & $239806$ & $358908$ & $354321$ \\ \hline
      Total $\#$ of Entities & $609$k & $8702$ & $278$k & $10684$ \\ \hline
      Total $\#$ of Predicates & $450$ & $256$ & $378$ & $395$ \\ \hline
      Avg. $\#$ of Triples (incl. Encoded Dates) per Article & $10.68$ & $10.68$ & $12.09$ & $11.96$ \\ \hline
      Max. $\#$ of Alloc. Triples (incl. Encoded Dates) per Article & $175$ & $22$ & $255$ & $21$ \\ \hline
      Total $\#$ of Words In the Summaries & $400$k & $14297$ & $500$k & $16728$ \\ \hline
      Total $\#$ of Annotated Entities In the Summaries & $194$k & $15703$ & $222$k & $16272$ \\ \hline
    \end{tabular}
    \label{table:DatasetsStatistics}
  \end{center}
\end{table*}

\subsubsection{Modelling the Input Triples}

Similar to the Wikipedia summaries, we represent the occurrence of the main discussed entity of the corresponding summary as either subject or object of a triple with the special \texttt{<item>} token. A shared, fixed dictionary was used for all subjects, predicates and objects. First, we included all the predicates and entities that occur at least $20$ times. Triples with rare predicates are discarded. Every out-of-vocabulary entity is replaced by the special token of its instance type, which is retrieved from the Instance Types dataset. For example the rare entity of \texttt{dbr:Mamma\_Mia!} is replaced by the \texttt{dbo:Musical} token. In case an infrequent entity is not found in the Instance Types dataset, it is replaced with the special \texttt{<unk>} token. We appended to the source vocabulary only the instance type tokens that occur at least $20$ times, and, finally, we used the \texttt{<resource>} token for the rare entities with also infrequent instance types.

In order to increase the homogeneity of the dataset in terms of the number of triples that are aligned with each Wikipedia summary, we limit the number of allocated triples per summary $E$ to:

\begin{align}
  \left \lfloor{E_{\min} + 0.25 \sigma_{E}}\right \rfloor \leq E \leq \left \lfloor{\overline{E} + 1.5 \sigma_{E}}\right \rfloor\enspace.
\end{align}In case a biography is aligned with less triples then it is excluded for the respective dataset. If a summary is aligned with more triples, we first attempt to exclude potential duplicates (e.g. \texttt{Fiorenzo\_Magni dbp:proyears 1945} and \texttt{Fiorenzo\_Magni dbp:proyears 1944} would result in the same triple: \texttt{<item> dbp:proyears <year>}). In case their number still exceeds the limit, we retain only the first ones until the threshold is reached.

Table \ref{table:DatasetsStatistics} shows statistics on the initial and the training-ready version of each corpus. An example of the structure of the datasets is displayed in Table \ref{table:TriplesSummariesDatasetAlignment}. More details about the two different types of summaries (i.e. Summary With URIs and Surface Form tuples) with which we trained our models are provided in Section \ref{sec:Experiments}.

\begin{table*}[h]
  \caption{Example of the alignment of our dataset. One Wikipedia summary is coupled with a set of triples from either DBpedia or Wikidata. Any reference to the main discussed entity of the summary (i.e.{\tt dbr:Papa\_Roach} or {\tt wikidata:Q254371} respectively) is replaced by the special \texttt{<item>} token both in the text and the corresponding triples. Each other entity is stored along with its instance type. In the case of infrequent entities these are replaced with the special token of their instance types both in the text and the triples (e.g. ``triple platinum'' is replaced with {\tt dbr:RIAA\_certification}). When a rare entity in the text is matched to an entity of the corresponding triples' set, then it is replaced by a unique token, which consists from the predicate of the triple, a descriptor of the component of the triple that was matched, and the instance type of the entity (e.g. the reference to the music album ``Infest (2000)'' is replaced with the placeholder \texttt{[dbo:artist\_\_sub\_\_dbo:Album]}).}
  
  \begin{center}
    \footnotesize
    \setlength{\extrarowheight}{1.5pt}
    \begin{tabular}{|R{3cm}||L{14cm}|}
      \hline
      {\tt<item>} & {\tt dbr:Papa\_Roach} and {\tt wikidata:Q254371} \\ \hline
      \textbf{Original Wikipedia Summary} & {Papa Roach is an American rock band from Vacaville, California. Formed in $1993$, their first major-label release was the triple-platinum album Infest ($2000$).} \\ \hline
      \multirow{8}{1.5cm}{\raggedleft\textbf{DBpedia Triples}} & {\tt<item> dbo:bandMember dbr:Jacoby\_Shaddix [dbo:MusicalArtist]} \\
                  & {\tt<item> dbo:bandMember dbr:Jerry\_Horton [dbo:MusicalArtist]} \\
                  & {\tt<item> dbo:genre dbr:Hard\_rock [dbo:MusicGenre]} \\
                  & \multicolumn{1}{c|}{\vdots} \\
                  & {\tt<item> dbo:hometown dbr:United\_States [dbo:Country]} \\
                  & {\tt<item> dbo:hometown dbr:Vacaville,\_California [dbo:City]} \\
      
                  & {\tt [dbo:Album] dbr:Infest\_(album) dbo:artist <item>} \\
                  & {\tt [dbo:Album] dbr:Metamorphosis\_(Papa\_Roach\_album) dbo:artist <item>} \\ \hline

      \textbf{Summary $/$w URIs} & {\texttt{<start>} \texttt{<item>} is an \texttt{dbr:United\_States} \texttt{dbr:Rock\_music} band from \texttt{[dbo:hometown\_\_obj\_\_dbo:City]} . Formed in \texttt{<year>} , their first major-label release was the \texttt{dbr:RIAA\_certification} album \texttt{[dbo:artist\_\_sub\_\_dbo:Album]} ( \texttt{<year>} ) . \texttt{<end>}} \\ \hline

      \textbf{Summary $/$w Surface Form Tuples} & {\texttt{<start>} \texttt{<item>} is an \texttt{(dbr:United\_States, American)} \texttt{(dbr:Rock\_music, rock)} band from \texttt{[dbo:hometown\_\_obj\_\_dbo:City]} . Formed in \texttt{<year>} , their first major-label release was the \texttt{dbr:RIAA\_certification} album \texttt{[dbo:artist\_\_sub\_\_dbo:Album]} ( \texttt{<year>} ) . \texttt{<end>}} \\ \hline\hline

      \multirow{8}{1.5cm}{\raggedleft \textbf{Wikidata Triples}} & {\tt<item> wikidata:P136 wikidata:Q11399 [dbo:MusicGenre]} \\
                  & {\tt<item> wikidata:P495 wikidata:Q30 [dbo:Country]} \\
                  & {\tt<item> wikidata:P571Month 1 [<unk>]} \\
                  & {\tt<item> wikidata:P571Year <year> [<unk>]} \\
                  & {\tt<item> wikidata:P31 wikidata:Q215380 [<unk>]} \\
                  & {\tt<item> wikidata:P264 wikidata:Q212699 [dbo:RecordLabel]} \\
                  & \multicolumn{1}{c|}{\vdots} \\
                  & {\tt [dbo:Album] wikidata:Q902353 wikidata:P175 <item>} \\ \hline

      \textbf{Summary $/$w URIs} & {\texttt{<start>} \texttt{<item>} is an \texttt{wikidata:Q30} \texttt{wikidata:Q11399} band from \texttt{dbo:City} . Formed in \texttt{<year>} , their first \texttt{<rare>} release was the \texttt{wikidata:Q2503026} album \texttt{[wikidata:P175\_\_sub\_\_dbo:Album]} ( \texttt{<year>} ) . \texttt{<end>}} \\ \hline
      
      \textbf{Summary $/$w Surface Form Tuples} & {\texttt{<start>} \texttt{<item>} is an \texttt{(wikidata:Q30, American)} \texttt{(wikidata:Q11399, rock)} band from \texttt{dbo:City} . Formed in \texttt{<year>} , their first \texttt{<rare>} release was the \texttt{<unk>} album \texttt{[wikidata:P175\_\_sub\_\_dbo:Album]} ( \texttt{<year>} ) . \texttt{<end>}} \\ \hline

    \end{tabular}
    \label{table:TriplesSummariesDatasetAlignment}
  \end{center}
  
\end{table*}

\section{Experiments}
\label{sec:Experiments}
We use the above datasets of aligned Wikipedia summaries with DBpedia and Wikidata triples in order to train and evaluate the performance of our neural network models. Both datasets are split into training, validation and test with respective portions of $85\%$, $10\%$, and $5\%$. The implemented architectures have been developed using the Torch\footnote{\href{http://torch.ch}{Torch} is a scientific computing package for Lua. It is based on the \href{http://luajit.org/}{LuaJIT} package.} software package. Any cleaning or restructuring procedure that has been carried out on the datasets has been conducted with Python scripts. The code will be made publicly available with the notification of acceptance.

Our proposed neural network architectures learn to generate a textual summary as a sequence of words and entities. In order to infer the verbalisation of the predicted entities in a generated summary, we experiment with two different approaches which are described in detail below.

\subsection{Generating Words Along With URIs}
\label{subsec:GenerateWithURIs}
In this setup, all the entities that have been annotated in the text with DBpedia Spotlight are replaced with their respective URIs. The summaries vocabulary is comprised of words and the entities' URIs. The model thus learns to generates words along with the URIs of entities. In order to improve the generated text further, as a post-processing step we replace:\begin{inparaenum}[(i)]\item the \texttt{<item>} token, with its corresponding surface form, and \item tokens of DBpedia or Wikidata entities in the text, with their most frequently matched surface form, as these are recorded during our data pre-processing (i.e. Section \ref{subsec:WikipediaSummaries})\end{inparaenum}.

\subsection{Generating Words Along With Surface Form Tuples}
\label{subsec:GenerateWithSurfaceFormTuples}
In order to eliminate the post-processing step of replacing the entities' URIs with their most frequently met surface forms, we propose a setup that enables our system to make a prediction about the best verbalisation of a predicted entity in the text. Each entity that has been identified in the text of the Wikipedia summaries using DBpedia Spotlight, is stored as a tuple of the annotated surface form and its URI. Let $K = \{k_1,k_2, \ldots, k_D\}$ be the set of all the $D$ entities that are annotated in the text. We define the $r$-th surface form tuple of the $d$-th entity $k_d$ as: $u_{r}^{k_d}=(k_d, g_r):k_d \in K$ , where $g_r$ is the $r$-th surface form that is associated with the entity $k_d$. Similarly to Section \ref{subsec:GenerateWithURIs}, those tuples are stored as tokens in the target vocabulary. This setup enables the models to verbalise each entity with more than one way by adapting the surface forms to the context of both the generated tokens and input triples.

\subsection{Training Details}

We train two different models. The first one is the triple encoder coupled with the GRU-based decoder to which we refer as Triples2GRU; the other is the same triple encoder coupled with the LSTM-based decoder (Triples2\-LSTM). For each dataset, we train each model on our task of generating a summary once as a combination of words with URIs ($/$w URIs, \ref{subsec:GenerateWithURIs}) and once as mixture of words and surface form tuples ($/$w Surf. Form Tuples, \ref{subsec:GenerateWithSurfaceFormTuples}).

\begin{table*}[h]
  \caption{Model Hyperparameters}
  \begin{center}
    \footnotesize
    {
      \setlength{\extrarowheight}{1.5pt}
      \begin{tabular}{|L{3.2cm}|C{1.3cm}|C{1.3cm}|C{1.3cm}|C{1.3cm}|C{1.3cm}|C{1.3cm}|C{1.3cm}|C{1.3cm}|} \hline
        \multirow{2}{*}{\textbf{Parameter}} & \multicolumn{2}{c|}{Triples2LSTM w$/$ URIs} & \multicolumn{2}{c|}{Triples2GRU w$/$ URIs} & \multicolumn{2}{C{3cm}|}{Triples2LSTM w$/$ Surf. Form Tuples} & \multicolumn{2}{C{3cm}|}{Triples2GRU w$/$ Surf. Form Tuples} \\ \cline{2-9}
                                            & DBpedia & Wikidata & DBpedia & Wikidata & DBpedia & Wikidata & DBpedia & Wikidata \\ \hline
        Batch Size & $85$ & $85$ &$85$ & $85$ & $85$ & $85$ & $85$ & $85$ \\ \hline
        Max. Timestep $T$ & $66$ &$69$ & $66$ & $69$ & $66$ & $69$ & $66$ & $68$ \\ \hline
        Embedding Size $m$ & $650$ & $650$ & $750$ & $750$ & $650$ & $650$ & $750$ & $750$ \\ \hline
        Target Vocabulary Size $|X|$ & $30692$ & $33644$ & $30692$ & $33644$ & $30761$ & $33715$ &  $30761$ & $33715$ \\ \hline
        Source Vocabulary Size $|N|$ & $9168$ & $11088$ & $9168$ & $11088$ & $9168$ & $11088$ & $9168$ & $11088$ \\ \hline
        Max. $\#$ of Alloc. Triples per Article $E_{\max}$ & $22$ & $21$  & $22$ & $21$  & $22$ & $21$  & $22$ & $21$ \\ \hline
        $\#$ Training Epochs\footnote{The epoch at which the model converges to the lowest possible validation error. After this epoch, either does not improve further or it increases (i.e. the model is overfitting).} & $12$ & $16$ &$12$ & $16$ & $12$ & $15$ & $13$ & $22$ \\ \hline
      \end{tabular}
    }
    \label{table:HyperParameters}
  \end{center}
\end{table*}

For the recurrent component of our networks, we use $1$ layer of\begin{inparaenum}[(i)]\item $650$ LSTM cells and \item $750$ GRUs\end{inparaenum}, resulting in $3.38$M and $3.375$M recurrent connections respectively. We found that increasing the number of layers does not improve the performance of our architectures, whereas the dimensionality of the hidden states plays a crucial role in achieving the best possible results. Table \ref{table:HyperParameters} summarises the hyper-parameters that have been used for the training of our models.

The feed-forward triples encoder consist of a sequence of fully-connected layers with the following $[\text{input}, \text{output}]$ sizes: \begin{inparaenum}[(i)]\item one-hot input to vector representation of subject or predicate or object: $[|N| \cdot m, m]$, \item concatenated vector representation of each triple subject-predicate-object to hidden representation of triple: $[3 \cdot m, m]$\end{inparaenum}. At the topmost layer of the encoder, we have a fully-connected layer that maps the concatenated hidden representations of all the aligned to a summary triples to one single vector: $[E_{\max} \cdot m, m]$, where $E_{\max}$ is the maximum number of triples per article. Sets of triples with less than $E_{\max}$ triples are \textit{padded} with zero vectors when necessary.

We optimised our architectures using an alteration of stochastic gradient descent with adaptive learning rate. We found that a fixed learning rate was resulting in the explosion of the gradients that were propagated to the encoder side of our models. We believe that the above behaviour is explained by the fact that our models learn to project data of dissimilar nature (i.e. structured data from the triples and unstructured text from the summaries) in a shared continuous semantic space. In case their parameters are not initialised properly, our neural architectures propagate vectors of different orders of magnitude leading to the explosion of the gradients phenomenon. However, finding the appropriate values to initialise the models' parameters is certainly not trivial \cite{Ioffe2015}. In order to sidestep this problem, we use Batch Normalisation before each non-linear activation function and after each fully-connected layer both on the encoder and the decoder side, and we initialise all parameters with random uniform distribution between $-0.001$ and $0.001$ \cite{Ioffe2015}. The networks were trained with mini-batch RMSProp with an initial learning rate value of $0.002$. Each update is computed using a mini-batch of $85$ dataset instances. An $l_2$ regularisation term over the parameters is also included in the cost function. After the $3$nd epoch, the learning rate was decayed by $0.8$ every half epoch.

We trained all of our models on a single Titan X (Pascal). The LSTM-based models complete an epoch of training: \begin{inparaenum}[(i)]\item in around $25$ minutes when trained on the Wikidata dataset, and \item $17$ minutes when trained on the DBpedia one\end{inparaenum}; the GRU-based architectures require \begin{inparaenum}[(i)]\item around $22$ minutes when trained on the Wikidata dataset, and \item $15$ minutes when trained on the DBpedia one\end{inparaenum}.

\begin{table*}[h]
  \caption{Automatic evaluation with the perplexity (i.e. lower is better), BLEU and ROUGE\textsubscript{L} metric (i.e. higher is better) on the validation and the test set. The average performance of the baseline along with its standard deviation is reported after sampling $10$ times.}
  \begin{center}
    \scriptsize
    {
      \setlength{\extrarowheight}{1.5pt}
      \begin{tabular}{|L{1.8cm}|C{.9cm}|C{.9cm}||C{.9cm}|C{.9cm}|C{.9cm}|C{.9cm}|C{.9cm}|C{.9cm}|C{.9cm}|C{.9cm}||C{.9cm}|C{.9cm}|} \hline
        \multirow{2}{*}{\textbf{Model}} & \multicolumn{2}{c||}{\textbf{Perplexity}} & \multicolumn{2}{c|}{\textbf{BLEU $1$}} & \multicolumn{2}{c|}{\textbf{BLEU $2$}} & \multicolumn{2}{c|}{\textbf{BLEU $3$}} & \multicolumn{2}{c||}{\textbf{BLEU $4$}} & \multicolumn{2}{c|}{\textbf{ROUGE\textsubscript{L}}}\\ \cline{2-13}
                                        & Valid. & Test & Valid. & Test & Valid. & Test & Valid. & Test & Valid. & Test & Valid. & Test \\ \hline
        
        Random Baseline on DBpedia & $-$ & $-$ & $29.523$ ($\pm .04$) & $29.650$ ($\pm .06$) & $17.270$ ($\pm .04$) & $17.390$ ($\pm .05$) & $11.415$ ($\pm .04$) & $11.528$ ($\pm .04$) & $7.561$ ($\pm .03$) & $7.658$ ($\pm .03$) & $27.578$ ($\pm .05$) & $27.715$ ($\pm .06$) \\ \hline
        KN on DBpedia & $-$ & $-$ & $22.587$ ($\pm .00$) & $22.685$ ($\pm .01$) & $16.601$ ($\pm .01$) & $16.722$ ($\pm .01$) & $12.626$ ($\pm .01$) & $12.750$ ($\pm .01$) & $9.412$ ($\pm .01$) & $9.518$ ($\pm .01$) & $38.202$ ($\pm .01$) & $38.418$ ($\pm .01$) \\ \hline
        Triples2LSTM on DBpedia $/$w URIs & $19.447$ & $19.769$ & $40.134$ & $39.902$ & $30.610$ & $30.430$ & $25.188$ & $25.025$ & $21.285$ & $21.121$ & $45.981$ & $45.937$ \\ \hline
        Triples2GRU on DBpedia w$/$ URIs & $20.530$ & $20.929$ & $41.003$ & $40.954$ & $31.557$ & $31.479$ & $26.088$ & $25.984$ & $22.116$ & $22.001$ & $\mathbf{47.092}$ & $47.100$ \\ \hline
        Triples2LSTM on DBpedia w$/$ Surf. Form Tuples & $19.171$ & $19.086$ & $40.679$ & $40.763$ & $30.809$ & $30.904$ & $25.234$ & $25.344$ & $21.287$ & $21.393$ & $44.973$ & $45.143$ \\ \hline
        Triples2GRU on DBpedia w$/$ Surf. Form Tuples & $20.164$ & $20.007$ & $\mathbf{41.350}$ & $\mathbf{41.457}$ & $\mathbf{31.877}$ & $\mathbf{31.991}$ & $\mathbf{26.387}$ & $\mathbf{26.510}$ & $\mathbf{22.419}$ & $\mathbf{22.531}$ & $47.027$ & $\mathbf{47.235}$  \\ \hline\hline

        Random Baseline on Wikidata & $-$ & $-$ & $29.636$ ($\pm .03$) & $29.650$ ($\pm .03$) & $17.587$ ($\pm .03$) & $17.581$ ($\pm .03$) & $11.818$ ($\pm .02$) & $11.800$ ($\pm .03$) & $7.910$ ($\pm .02$) & $7.892$ ($\pm .03$) & $28.083$ ($\pm .04$) & $28.109$ ($\pm .04$) \\ \hline
        KN on Wikidata & $-$ & $-$ & $22.716$ ($\pm .00$) & $22.713$ ($\pm .00$) & $16.680$ ($\pm .00$) & $16.675$ ($\pm .00$) & $12.692$ ($\pm .00$) & $12.685$ ($\pm .00$) & $9.448$ ($\pm .01$) & $9.432$ ($\pm .01$) & $37.937$ ($\pm .00$) & $37.957$ ($\pm .01$) \\ \hline        
        Triples2LSTM on Wikidata w$/$ URIs & $20.995$ & $21.045$ & $40.967$ & $41.134$ & $31.190$ & $31.312$ & $25.729$ & $25.812$ & $21.780$ & $21.845$ & $46.522$ & $46.675$ \\ \hline
        Triples2GRU on Wikidata w$/$ URIs & $21.770$ & $21.823$ & $41.470$ & $\mathbf{41.618}$ & $31.920$ & $32.072$ & $26.475$ & $26.604$ & $22.497$ & $22.619$ & $47.577$ & $47.761$ \\ \hline
        Triples2LSTM on Wikidata w$/$ Surf. Form Tuples & $20.779$ & $20.403$ & $40.660$ & $40.604$ & $31.158$ & $31.144$ & $25.776$ & $25.783$ & $21.874$ & $21.898$ & $47.031$ & $47.140$ \\ \hline
        Triples2GRU on Wikidata w$/$ Surf. From Tuples  & $21.493$ & $21.200$ & $\mathbf{41.527}$ & $41.566$ & $\mathbf{32.072}$ & $\mathbf{32.097}$ & $\mathbf{26.645}$ & $\mathbf{26.673}$ & $\mathbf{22.679}$ & $\mathbf{22.708}$ & $\mathbf{47.979}$ & $\mathbf{48.100}$\\ \hline       
      \end{tabular}
    }
    \label{table:Performance}
  \end{center}
\end{table*}

\subsection{Automatic Evaluation}
\label{subsec:AutomaticEvaluation}

\begin{figure*}[t]
  \subfloat[][DBpedia\label{fig:complexity-p}]{\centering
  \def\svgwidth{0.475\linewidth}
  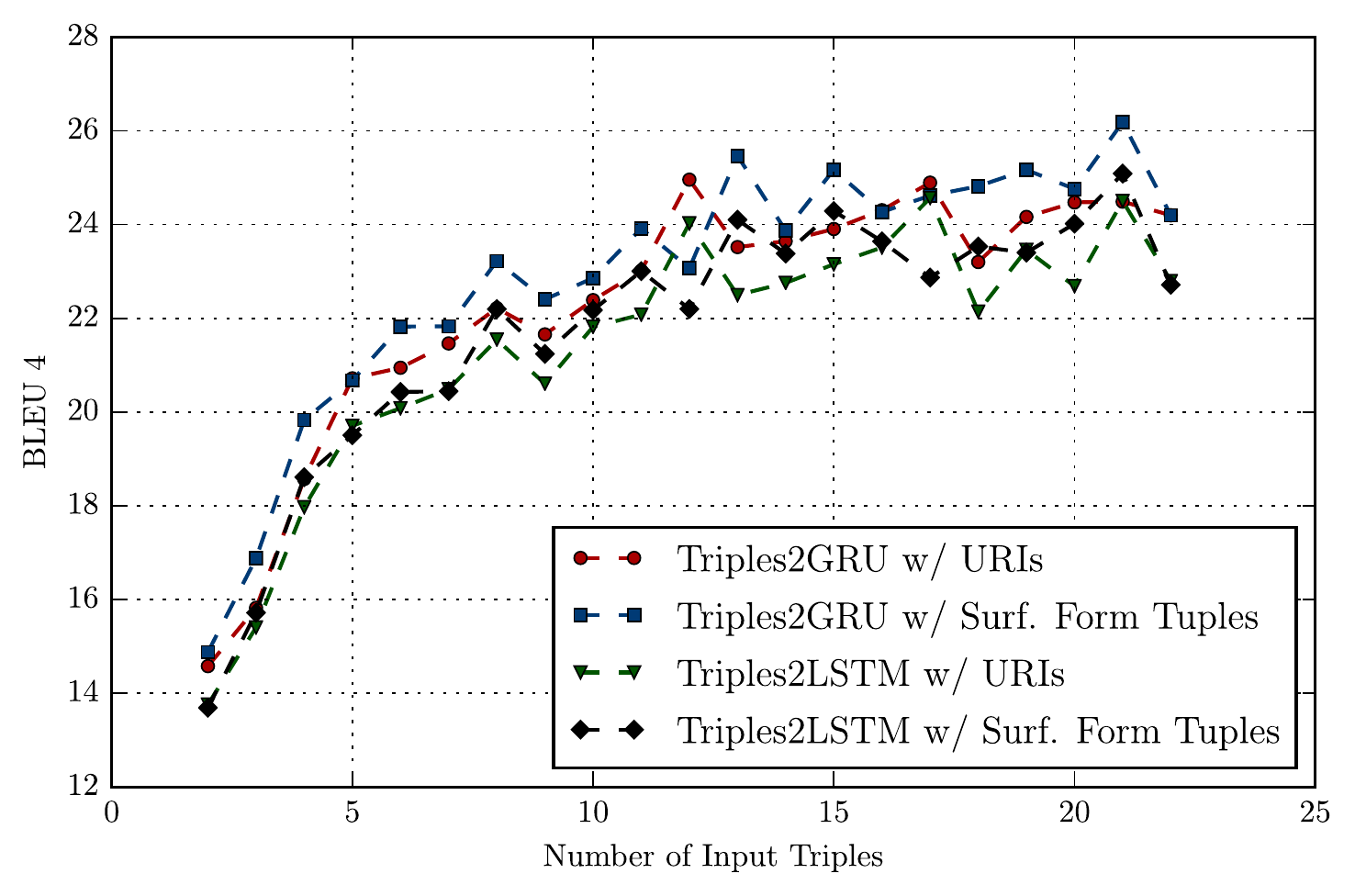}\hfill
  \subfloat[][Wikidata\label{fig:complexity-p}]{\centering
  \def\svgwidth{.475\linewidth}
  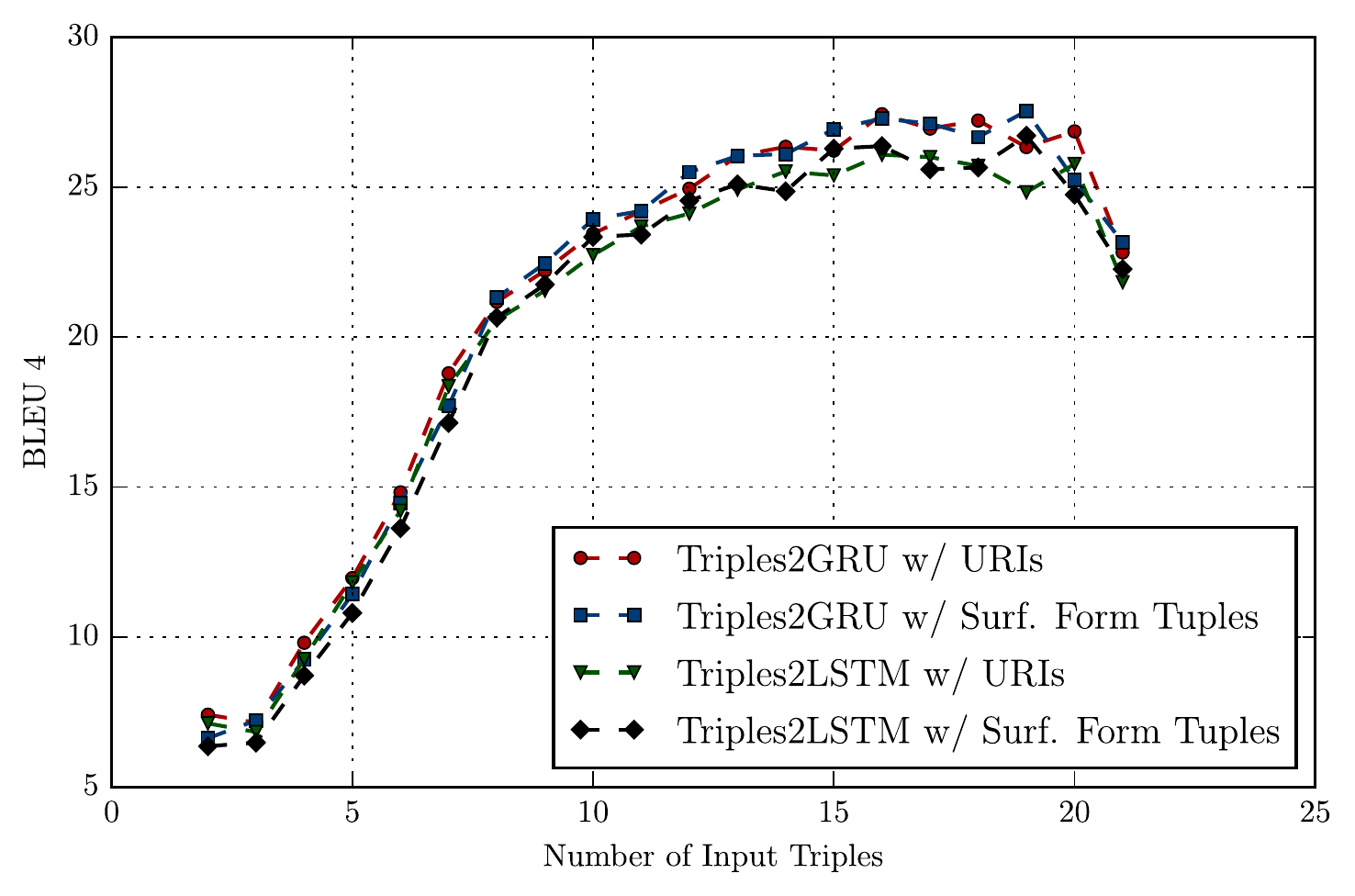}\hfill
\caption{Performance of our models with the BLEU $4$ metric across the difference number of input triples on DBpedia~(a) and Wikidata~(b).}
\label{fig:BLEU}
\end{figure*}

We use perplexity\footnote{Perplexity measures the cross-entropy between the predicted sequence of words and the actual, empirical, sequence of words.}, BLEU\footnote{BLEU (Bilingual Evaluation Understudy) \cite{Papineni2002} is precision-oriented metric for measuring the quality of generated text by comparing it to the actual, empirical text. BLEU-$n$ calculates a similarity scores based on the co-occurrence of up to $n$-grams in the generated and the actual text.}, and ROUGE\footnote{ROUGE (Recall-Oriented Understudy for Gisting Evaluation) is a metric that computes the recall of $n$-grams in the generated text with respect to the $n$-grams of the actual text \cite{Lin2004}.} on the validation and test set. Perplexity indicates how well the model learns its training object (i.e. Section \ref{subsec:GeneratingSummary}), whereas BLEU and ROUGE measure how close is the generated text to the actual Wikipedia summary. Essentially, BLEU and ROUGE are complimentary to each other. The first computes a modified version of $n$-grams precision\footnote{The count of True Positives of an $n$-gram in the generated summary has an upper bound which is defined by the number of occurrences of this particular $n$-gram in the actual summary.}, whereas the latter computes $n$-grams recall, of the automatically generated sentences with respect to the empirical Wikipedia summaries.

We adapt the code from the evaluation package that was released by Peter Anderson\footnote{\href{https://github.com/peteanderson80/coco-caption}{\texttt{http://github.com/peteanderson80/coco-caption}}}, which was originally implemented to score textual descriptions from images. Perplexity, BLEU $1$, BLEU $2$, BLEU $3$, BLEU $4$, and ROUGE\textsubscript{L} (i.e. an alteration of the original ROUGE that is automatically measured on the longest common sub-sequence) results are reported in Table \ref{table:Performance}.

To demonstrate the effectiveness of our system, we compare it to two baselines. First, we compute expected lower bounds for BLEU scores by using a random Wikipedia summary generation baseline. We regard the latter as a particularly strong baseline, due to the fact that Wikipedia biographies tend to follow a limited number of textual patterns. For each triples set on the validation and test set, the random system generates a response by randomly selecting a Wikipedia summary from our training set. Secondly, we use the KenLM toolkit \cite{Heafield2013} in order to build a $5$-gram Kneser-Ney (KN) language model. During testing, similarly to the case of our neural network approach, for each triple set in the validation and test set, we use beam-search with a beam of size $10$, in order to generate the $10$ most probable summaries. We equip both of our baselines with surface form tuples, and the \texttt{<item>} and property-type placeholders. After a summary is selected, its \texttt{<item>} placeholder along with any potential property-type placeholders are replaced based on the original triples. In case a property-type placeholder are not matched to the content of the triples they are replaced by their corresponding instance type token (i.e. Section \ref{subsubsec:ModellingGeneratedSummaries}). The results are illustrated in Table \ref{table:Performance}.

Our scores are lower than those usually reported for machine translation tasks. However, they should be indicative of how well can our model generate a Wikipedia summary given the set of its corresponding triples. Furthermore, it should be noted that our task consists of learning to generate text from a corpus of triples loosely associated with text which is not the case in machine translation where there is a tight alignment between the source and the generated language.

In addition to the above experiments, we group Wikipedia summaries that are allocated to same number of input triples and compute a BLEU score per group. Figure \ref{fig:BLEU} displays the performance of our models with the BLEU $4$ metric across different numbers of input triples. The low performance of the models when they are initialised with a low number of triples is explained by the fact that the systems are lacking information required to form a two-sentence summary.

\begin{table*}[h!]
  \caption{Examples of a textual summaries that are generated by our proposed systems given unknown sets of triples as input. For each model, we report each immediate output along with its corresponding (Final) version after the replacement of the generated placeholder tokens. Each other than the main discussed entity (\texttt{<item>}) in the triples is recorded and displayed along with its instance type. Rare entities in the input triples are replaced with their respective instance type tokens. The greyed-out tokens of either entities or instance type tokens refer to entities that are not used during the training of our neural net models.}
  \begin{center}
    \scriptsize
    \setlength{\extrarowheight}{1.5pt}
    \begin{tabular}{|L{3.5cm}|L{14cm}|}\hline

      {\tt<item>} & {\tt dbr:Barbara\_Flynn}  \\ \hline
      \multirow{10}{*}{\textbf{Triples}} & \texttt{<item> dbo:birthPlace dbr:England \textcolor{grey}{[owl\#Thing]}} \\
                  & \texttt{<item> dbo:birthPlace \textcolor{grey}{dbr:St\_Leonards-on-Sea} [dbo:Settlement]} \\
                  & \texttt{<item> dbo:birthPlace dbr:Sussex \textcolor{grey}{[owl\#Thing]}} \\
                  & \texttt{<item> dbo:occupation dbr:Actress \textcolor{grey}{[<unk>]}} \\
                  & \texttt{<item> dbo:birthDateMonth 8 \textcolor{grey}{[<unk>]}} \\
      
                  & \texttt{<item> dbo:birthDateYear <year> \textcolor{grey}{[<unk>]}} \\
                  & \texttt{[dbo:TelevisionShow] \textcolor{grey}{dbr:Open\_All\_Hours} dbo:starring <item>} \\
                  & \texttt{[dbo:TelevisionShow] \textcolor{grey}{dbr:A\_Very\_Peculiar\_Practice} dbo:starring <item>} \\
                  & \texttt{[dbo:TelevisionShow] \textcolor{grey}{dbr:The\_Beiderbecke\_Trilogy} dbo:starring <item>} \\
                  & \vdots \\
                  & \texttt{[dbo:TelevisionShow] \textcolor{grey}{dbr:Cracker\_(UK\_TV\_series)} dbo:starring <item>} \\ \hline
      
      \textbf{Triples2GRU w$/$ URIs} & \texttt{<start>} \texttt{<item>} ( born \texttt{0} August \texttt{<year>} ) is an \texttt{dbr:English\_people} \texttt{dbr:Actor} and \texttt{dbr:Actor}. She is best known for her roles in the \texttt{dbr:Television\_program} \texttt{[dbo:starring\_\_sub\_\_dbo:TelevisionShow]} , \texttt{[dbo:starring\_\_sub\_\_dbo:TelevisionShow]} and \texttt{[dbo:starring\_\_sub\_\_dbo:TelevisionShow]} . \texttt{<end>}\\ \hline
      
      \textbf{Triples2GRU w$/$ URIs (Final)} & \texttt{<start>} Barbara Flynn ( born \texttt{0} August \texttt{<year>} ) is an English actor and actor . She is best known for her roles in the television series A Very Peculiar Practice , Beiderbecke Trilogy and Open All Hours . \texttt{<end>} \\ \hline
      
      \textbf{Triples2GRU w$/$ Surf. Form Tuples} & \texttt{<start>} \texttt{<item>} ( born \texttt{0} August \texttt{<year>} ) is an \texttt{(dbr:English\_people, English)} \texttt{(dbr:Actor, actress)} . She is best known for her roles in \texttt{[dbo:starring\_\_sub\_\_dbo:TelevisionShow]} and \texttt{[dbo:starring\_\_sub\_\_dbo:TelevisionShow]} . \texttt{<end>} \\ \hline
      
      \textbf{Triples2GRU w$/$ Surf. Form Tuples (Final)} & \texttt{<start>} Barbara Flynn ( born \texttt{0} August \texttt{<year>} ) is an English actress . She is best known for her roles in Beiderbecke Trilogy and Open All Hours . \texttt{<end>} \\ \hline

      \textbf{Triples2LSTM w$/$ URIs} & \texttt{<start>} \texttt{<item>} ( born \texttt{0} August \texttt{<year>} ) is an \texttt{dbr:English\_people} \texttt{dbr:Actor} . She is best known for her role as \texttt{dbo:SoapCharacter} in the BBC soap opera EastEnders . \texttt{<end>} \\ \hline
      
      \textbf{Triples2LSTM w$/$ URIs (Final)} & \texttt{<start>} Barbara Flynn ( born \texttt{0} August \texttt{<year>} ) is an English actor .  She is best known for her role as \texttt{dbo:SoapCharacter} in the BBC soap opera EastEnders . \texttt{<end>} \\ \hline
      
      \textbf{Triples2LSTM w$/$ Surf. Form Tuples} & \texttt{<start>} \texttt{<item>} ( born \texttt{0} August \texttt{<year>} ) is an \texttt{(dbr:English\_people, English)} \texttt{(dbr:Actor, actress)} .  She is best known for her role as \texttt{dbo:SoapCharacter} in the \texttt{(dbr:BBC, BBC)} soap opera \texttt{(dbr:EastEnders, EastEnders)} . \texttt{<end>} \\ \hline
      
      \textbf{Triples2LSTM w$/$ Surf. Form Tuples (Final)} & \texttt{<start>} Barbara Flynn ( born \texttt{0} August \texttt{<year>} ) is an English actress . She is best known for her role as \texttt{dbo:SoapCharacter} in the BBC soap opera EastEnders . \texttt{<end>} \\ \hline\hline

      {\tt<item>} & {\tt dbr:Lee\_Jeong-beom}  \\ \hline
      \multirow{10}{*}{\textbf{Triples}} & \texttt{<item> dbo:birthPlace dbr:South\_Korea \textcolor{grey}{[dbo:Country]}} \\
                                                  & \texttt{<item> dbo:education  \textcolor{grey}{dbr:Korea\_National\_University\_of\_Arts} [dbo:University]} \\
                                                  & \texttt{<item> dbo:occupation dbr:Film\_director  \textcolor{grey}{[owl\#Thing]}} \\
                                                  & \texttt{<item> dbo:occupation dbr:Screenwriter  \textcolor{grey}{[owl\#Thing]}} \\
                                                  & \texttt{<item> dbo:birthDateMonth 9  \textcolor{grey}{[<unk>]}} \\
                                                  & \texttt{<item> dbo:birthDateYear <year>  \textcolor{grey}{[<unk>]}} \\
                                                  & \vdots \\
                                                  & \texttt{[dbo:Film]  \textcolor{grey}{dbr:Cruel\_Winter\_Blues} dbo:director <item>} \\
                                                  & \texttt{[dbo:Film]  \textcolor{grey}{dbr:Cruel\_Winter\_Blues} dbo:writer <item>} \\ \hline

      \textbf{Triples2GRU w$/$ URIs} &  \texttt{<start>} \texttt{<item>} (born September \texttt{0} , \texttt{<year>} ) is a \texttt{dbr:South\_Korea} \texttt{dbr:Film\_director} and \texttt{dbr:Screenwriter} . He is best known for his films \texttt{[dbo:director\_\_sub\_\_dbo:Film]} ( \texttt{<year>} ) and \texttt{[dbo:director\_\_sub\_\_dbo:Film]} ( \texttt{<year>} ) . \texttt{<end>} \\ \hline
      
      \textbf{Triples2GRU w$/$ URIs (Final)} & \texttt{<start>} Lee Jeong-beom ( born September \texttt{0} , \texttt{<year>} ) is a South Korean film director and screenwriter . He is best known for his films Cruel Winter Blues ( \texttt{<year>} ) and \texttt{dbo:Film} ( \texttt{<year>} ) . \texttt{<end>} \\ \hline

      \textbf{Triples2GRU w$/$  Surf. Form Tuples} & \texttt{<start>} \texttt{<item>} ( \texttt{[dbr:Hangul, Hangul]} : <rare> ; born September \texttt{0} , \texttt{<year>} ) is a \texttt{(dbr:South\_Korea, South Korean)} \texttt{(dbr:Film\_director, film director)} and \texttt{(dbr:Screenwriter, screenwriter)} . He is best known for directing the \texttt{<year>} \texttt{(dbr:Film\_director, film)} \texttt{[dbo:director\_\_sub\_\_dbo:Film]} . \texttt{<end>} \\ \hline

      \textbf{Triples2GRU w$/$ Surf. Form Tuples (Final)} & \texttt{<start>} Lee Jeong-beom ( Hangul : <rare> ; born September \texttt{0} , \texttt{<year>} ) is a South Korean film director and screenwriter . He is best known for directing the \texttt{<year>} film Cruel Winter Blues . \texttt{<end>} \\ \hline

      \textbf{Triples2LSTM w$/$ URIs} &  \texttt{<start>} \texttt{<item>} (born September \texttt{0} , \texttt{<year>} ) is a \texttt{dbr:South\_Korea} \texttt{dbr:Film\_director} and \texttt{dbr:Screenwriter} . He has directed more than \texttt{0} films since \texttt{year} . \texttt{<end>} \\ \hline
      
      \textbf{Triples2LSTM w$/$ URIs (Final)} & \texttt{<start>} Lee Jeong-beom ( born September \texttt{0} , \texttt{<year>} ) is a South Korean film director and screenwriter . He has directed more than \texttt{0} films since \texttt{<year>} . \texttt{<end>} \\ \hline

      \textbf{Triples2LSTM w$/$ Surf. Form Tuples} & \texttt{<start>} \texttt{<item>} ( born September \texttt{0} , \texttt{<year>} ) is a \texttt{(dbr:South\_Korea, South Korean)} \texttt{(dbr:Film\_director, film director)} and \texttt{(dbr:Screenwriter, screenwriter)} . He has directed \texttt{0} films since \texttt{<year>} . \texttt{<end>} \\ \hline

      \textbf{Triples2LSTM w$/$ Surf. Form Tuples (Final)} & \texttt{<start>} Lee Jeong-beom ( born September \texttt{0} , \texttt{<year>} ) is a South Korean film director and screenwriter . He has directed \texttt{0} films since \texttt{<year>} . \texttt{<end>} \\ \hline

    \end{tabular}
    
    \label{table:TextGenerationExamples}
  \end{center}
\end{table*}

\subsection{Human Evaluation}

Given the exploratory nature of our task, human evaluation is necessary in order to objectively assess the performance of our approach. The human evaluation was conducted using seven researchers, all of whom are experts in the field of the Semantic Web. For each corpora, we compiled a list of $15$ randomly selected sets of triples along with the textual summaries that have been generated from each one of our proposed models (i.e.\begin{inparaenum}[(i)]\item GRU with URIs and surface forms, and \item LSTM with URIs and surface forms\end{inparaenum}). The sets of triples are sampled from the test set. We conducted two separate experiments, one for each corpora.

\begin{table*}[t]
  \caption{The average rating of our models against the human evaluation criteria. For fluency and summarised triples the higher the score the better; for contradicting triples and additional information, the lower the score the better. The results are reported in the ``mean $\pm$ standard deviation'' format.} 
  \begin{center}
    \footnotesize
    {
    \setlength{\extrarowheight}{1.5pt}
    \begin{tabular}{|L{3cm}|C{2.55cm}|C{2.55cm}|C{2.55cm}|C{2.55cm}|}\hline
      \textbf{Model} & \textbf{Fluency} & \textbf{Summarised Triples} & \textbf{Contradicting Triples} & \textbf{Additional Information} \\ \hline 
      Triples2LSTM on DBpedia w$/$ URIs & $5.124$ ($\pm 0.963$) & $0.4$ ($\pm 0.169$) & $0.045$ ($\pm 0.069$) & $0.143$ ($\pm 0.151$) \\ \hline
      Triples2LSTM on DBpedia w$/$ Surf. Form Tuples & $5.287$ ($\pm 0.791$) & $0.457$ ($\pm 0.236$) & $0.05$ ($\pm 0.068$) & $0.145$ ($\pm 0.169$) \\ \hline
      Triples2GRU on DBpedia w$/$ URIs & $4.9$ ($\pm 1.006$) & $0.423$ ($\pm 0.221$) & $0.023$ ($\pm 0.06$) & $\mathbf{0.112}$ ($\pm 0.141$) \\ \hline
      Triples2GRU on DBpedia w$/$ Surf. Form Tuples & $\mathbf{5.511}$ ($\pm 0.640$) & $\mathbf{0.497}$ ($\pm 0.247$) & $\mathbf{0.017}$ ($\pm 0.056$) & $0.134$ ($\pm 0.177$) \\ \hline\hline

      Triples2LSTM on Wikidata w$/$ URIs & $5.036$ ($\pm 1.017$) & $0.582$ ($\pm 0.185$) & $0.018$ ($\pm 0.037$) & $0.103$ ($\pm 0.109$) \\ \hline
      Triples2LSTM on Wikidata w$/$ Surf. Form Tuples & $5.470$ ($\pm 0.687$) & $0.582$ ($\pm 0.185$) & $0.018$ ($\pm 0.037$) & $0.103$ ($\pm 0.109$) \\ \hline
      Triples2GRU on Wikidata w$/$ URIs & $5.349$ ($\pm 0.833$) & $0.596$ ($\pm 0.200$) & $\mathbf{0.006}$ ($\pm 0.023$) & $0.085$ ($\pm 0.107$) \\ \hline
      Triples2GRU on Wikidata w$/$ Surf. Form Tuples & $\mathbf{5.663}$ ($\pm 0.668$) & $\mathbf{0.597}$ ($\pm 0.194$) & $0.009$ ($\pm 0.028$) & $\mathbf{0.073}$ ($\pm 0.101$) \\ \hline
    \end{tabular}
    }
    \label{table:HumanEvaluation}
  \end{center}
\end{table*}

\begin{table*}[h!]
  \caption{Nearest neighbours of the vector representations of some of the most frequently occurring entities as these are learned by the encoder.}
  \begin{center}
    \scriptsize
    \setlength{\extrarowheight}{1.5pt}
    \begin{tabular}{|L{2.7cm}|L{11.5cm}|}\hline

      {\textbf{DBpedia Entity}} & \textbf{Nearest Neighbours} \\ \hline
      {\texttt{dbr:France}} & \texttt{dbr:Paris,\_France}, \texttt{dbr:Marseille}, \texttt{dbr:Lyon}, \texttt{dbr:Kingdom\_of\_France}, and \texttt{dbr:Olympique\_de\_Marseille} \\ \hline
      {\texttt{dbr:Japan}} & \texttt{dbr:Empire\_of\_Japan}, \texttt{dbr:Chiba\_Prefecture}, \texttt{dbr:Yokohama}, \texttt{dbr:Osaka}, and \texttt{dbr:Kyoto} \\ \hline
      {\texttt{dbr:Singer}} & \texttt{dbr:Singing}, \texttt{dbr:Vocalist}, \texttt{dbr:Vocals}, \texttt{dbr:Playback\_singer}, and \texttt{dbr:Americana\_(music)} \\ \hline
      {\texttt{dbr:Heavy\_metal\_music}} & \texttt{dbr:Glam\_metal}, \texttt{dbr:Doom\_metal}, \texttt{dbr:Hard\_rock}, \texttt{dbr:Nu\_metal}, and \texttt{dbr:Alternative\_metal} \\ \hline
      {\texttt{dbr:FC\_Barcelona}} & \texttt{dbr:RCD\_Mallorca}, \texttt{dbr:Athletic\_Bilbao}, \texttt{dbr:Spain\_national\_under-18\_football\_team}, \texttt{dbr:Valencia\_CF}, and \texttt{dbr:Battle\_of\_the\_Atlantic} \\ \hline
      \multicolumn{2}{c}{} \\ \hline
      
      {\textbf{Wikidata Entity}} & \textbf{Nearest Neighbours} \\ \hline
      {\texttt{wikidata:Q64}} (Berlin) & \texttt{wikidata:Q1022} (Stuttgart), \texttt{wikidata:Q365} (Taiwan), \texttt{wikidata:Q152087} (Humboldt University of Berlin), \texttt{wikidata:Q1731} (Dresden), and \texttt{wikidata:Q43287} (German Empire) \\ \hline

      {\texttt{wikidata:Q148}} (China) & \texttt{wikidata:Q17427} (Communist Party of China), \texttt{wikidata:Q865} (Taiwan), \texttt{wikidata:Q7850} (Chinese language), \texttt{wikidata:Q8686} (Shanghai), and \texttt{wikidata:Q1348} (Kolkata) \\ \hline

      {\texttt{wikidata:Q20}} (Norway) & \texttt{wikidata:Q35} (Denmark), \texttt{wikidata:Q486156} (University of Oslo), \texttt{wikidata:Q9043} (Norwegian language), \texttt{wikidata:Q11739} (Lahore), and \texttt{wikidata:Q585} (Oslo) \\ \hline

      {\texttt{wikidata:Q15981151}} (jazz musician) & \texttt{wikidata:Q12800682} (saxophonist), \texttt{wikidata:Q248970} (Berklee College of Music), \texttt{wikidata:Q806349} (bandleader), \texttt{wikidata:Q12804204} (percussionist), and \texttt{wikidata:Q8341} (jazz) \\ \hline
      
       {\texttt{wikidata:Q158852}} (conductor) & \texttt{wikidata:Q1415090} (film score composer), \texttt{wikidata:Q9734} (symphony), \texttt{wikidata:Q3455803} (director), \texttt{wikidata:Q1198887} (music director), and \texttt{wikidata:Q2994538} (Conservatoire national sup\'{e}rieur de musique et de danse) \\ \hline
    
    \end{tabular}
    
    \label{table:NearestNeighbours}
  \end{center}
\end{table*}

Our experiment showed that in our dataset, sets with less triples usually lack enough information for our model to generate a summary (i.e. Section \ref{subsec:AutomaticEvaluation}). Hence, we included only sets that consist of at least $6$ triples. The specific model to which each generated summary corresponds to (i.e. LSTM or GRU with URIs or surface form tuples) was anonymised. The evaluators were asked to rate each generated summary against four different criteria:\begin{inparaenum}[(i)]\item fluency\footnote{Adapted from \cite{NgongaNgomo2013} where they use an identical metric to evaluate the comprehensibility and readability of a generated question in natural language.}, \item number of contradicting facts (i.e. information that exists in the sentence but it conflicts with a number of triples from the input set), \item number of summarised triples\footnote{Similar to \textit{coverage} in \cite{Ell2014} which measures the number of included sub-graphs in the text.} (i.e. triples whose information is mentioned either implicitly or explicitly in the text), and \item number of triples to which potential additional information in the text can be interpreted\end{inparaenum}. We divide each score for the three latter criteria with the total number of triples of their respective set. Fluency was marked on a scale from $1$ to $6$, with $1$ indicating an incomprehensible summary, $2$ a barely understandable summary with significant grammatical errors, $3$ an understandable summary with grammatical flaws, $4$ a comprehensible summary with minor grammatical errors, $5$ a comprehensible and grammatically correct summary that reads a bit artificial, and $6$ a coherent and grammatically correct summary \cite{NgongaNgomo2013}.

The results of the human evaluators are aligned with the result of our automatic evaluation with both the BLEU and ROUGE metrics (i.e. Section \ref{subsec:AutomaticEvaluation}). The GRU-based architectures outperforms the LSTM-based in all criteria. Furthermore, they score consistently better in terms of the inclusion of additional or contradicting information. Since they are more reluctant to introduce out-of-context information in the text, their generated textual content is better aligned with the input triples.

In general the evaluators scored all of our models with high fluency ratings, thus, emphasising the ability of our approach to generate grammatically and syntactically correct text. We note, however, that verbalising the occurrence of entities in the text with the mechanism of surface form tuples makes all the investigated setups more fluent.

\subsection{Discussion}

Two examples of textual summaries that are generated by our models are shown in Table \ref{table:TextGenerationExamples}. We selected two representative sets of triples from the test set. The examples illustrate our approaches' capability of generating sentences that couple information from several triples from each set. In the first example, all the models are able to capture the main entity's gender for the input triple set. However, only in the case of the models equipped with surface form tuples, we are able to correctly verbalise the entity of \texttt{dbr:Actor} correctly as ``actress'' in the text. This is due to the fact that in the biographies dataset, the most frequent surface, with which the entity of \texttt{dbr:Actor} has been associated, is ``actor''. Subsequently, actor is used as the replacement of all the occurrences of the $dbr:Actor$ entity in the summaries that are generated by our $w/$ URIs models.

The learned embeddings on the decoder side capture information that is both coupled with the embeddings on the encoder side (e.g. the embedding of the pronouns ``She'' and ``her'' are coupled implicitly with the existence of the triple: \texttt{<item>} \texttt{dbo:occupation} \texttt{dbr:Actress}, and their own probability of occurring in the context of the sequentially generated text (e.g. word with its first letter capitalised when it is following a full stop). Consequently, items that have similar semantic meaning find themselves close together in the continuous semantic space. Table \ref{table:NearestNeighbours} shows the nearest neighbours of some of the most frequently occurring entities in our datasets which have been learned by our models. This illustrates our models' capability to successfully infer semantic relationships among entities.

The main drawback of training our models on a dataset of loosely associated triples with text is that the information that exists in the triples does not necessarily appear in corresponding text, and vice versa. As a result, the models are not penalised when they generates textual content that does not exist in the set of input triples.

\section{Conclusion}
\label{sec:Conclusion}

To the best of our knowledge this work constitutes the first attempt to use neural networks for Natural Language Generation on top of Semantic Web triples. We propose an end-to-end trainable system that is able to generate a textual summary given a set of triples as input. The generated summary discusses various aspects of the information encoded within the input triple set.

Our approach does not require any hand-engineered templates and can be applied to a great variety of domains. We propose an method of building a loosely aligned dataset of DBpedia and Wikidata triples with Wikipedia summaries in order to satisfy the training requirements of our system. Using these datasets, we have demonstrated that our technique is capable of scaling to domains with vocabularies of over $400$k words. We address the problem of learning high quality vector representations for rare entities by adapting a multi-placeholder approach that enables to lexicalise rare entities in the text. Our models learn to emit those placeholder tokens that are replacing with the surface form of the corresponding entities in the triples at a post-processing step.

We use a series of well-established automatic text similarity metrics in order to automatically evaluate our approach's ability of predicting the Wikipedia summary that corresponds to a set of unknown triples showing substantial improvement over our baselines. Furthermore, our choice to introduce a statistical approach for inferring the verbalisation of the entities in the text, further enhances the, reported by our human evaluators, fluency of the generated summaries compared to a purely deterministic replacement of the generated entities' URIs.

\bibliography{Bibliography}

\end{document}